\documentclass[acmtog]{acmart}

\AtBeginDocument{%
  \providecommand\BibTeX{{%
    \normalfont B\kern-0.5em{\scshape i\kern-0.25em b}\kern-0.8em\TeX}}}

\usepackage{soul,xcolor}

\usepackage{lipsum}
\usepackage{cleveref}
\usepackage{graphicx}

\usepackage{amsmath}

\begin{document}

\title{Solutions to Deepfakes: Can Camera Hardware, Cryptography, and Deep Learning Verify Real Images?}

\author{Alexander Vilesov, Yuan Tian, Nader Sehatbakhsh, Achuta Kadambi \\ University of California, Los Angeles (UCLA)}

\received{20 February 2024}
\received[revised]{20 February 2024}
\received[accepted]{20 February 2024}

\begin{teaserfigure}
    \centering
    \includegraphics[width=\textwidth]{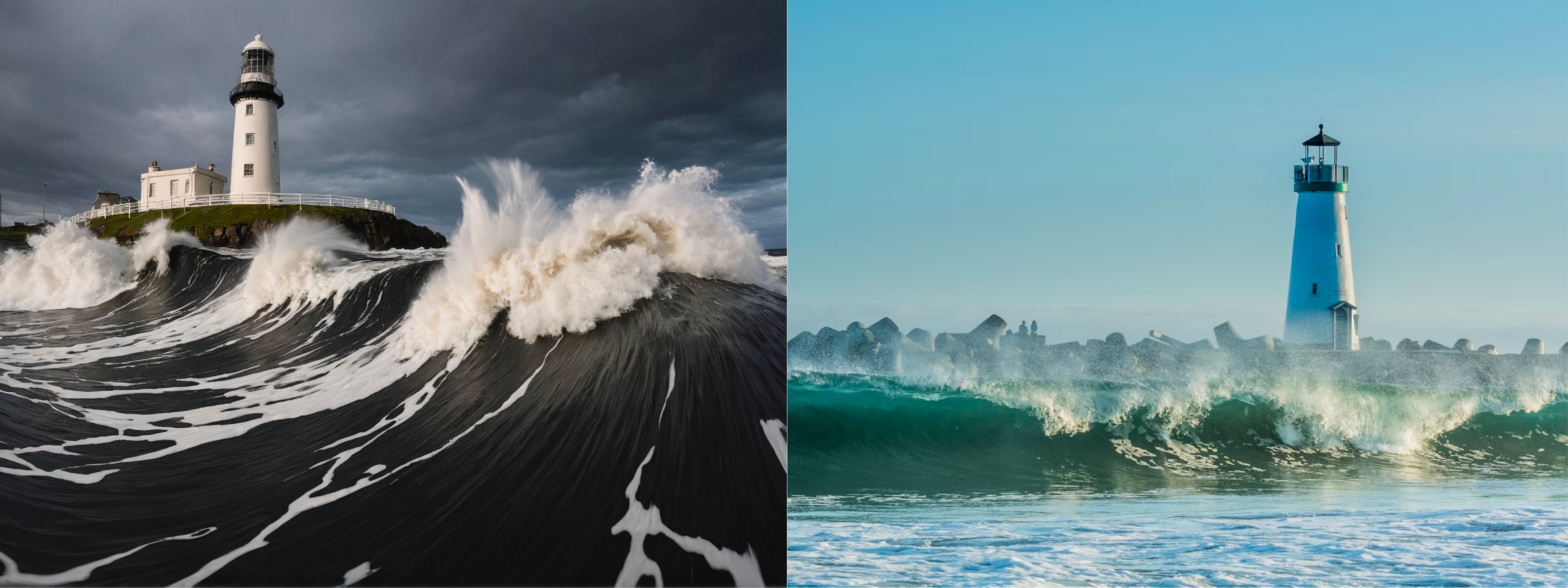}
    \caption{\textbf{Photo of a lighthouse generated by AI~\cite{SD3} on the left and a real photo on the right.} As synthetic image generators improve, the near future will see synthetic images being generated that are visually indistinguishable from real images. Distinguishing between real and synthetic is going to be extremely important and existing infrastructure is not sufficient.}
    \label{fig:enter-label}
\end{teaserfigure}

\maketitle

\section{Introduction}
The exponential progress in generative AI poses serious implications for the credibility of all real images and videos. There will exist a point in the future where 1) digital content produced by generative AI will be indistinguishable from those created by cameras, 2) high-quality generative algorithms will be accessible to anyone, and 3) the ratio of all synthetic to real images will be large. It is imperative to establish methods that can separate real data from synthetic data with high confidence. We define `real images' as those that were produced by the camera hardware, capturing a real-world scene. Any synthetic generation of an image or alteration of a real image through generative AI or computer graphics techniques is labeled as a `synthetic image'. To this end, this document aims to:
\begin{enumerate}
    \item present known strategies that can be employed to verify which images are `real'
    \item weight the strengths and weaknesses of these strategies
    \item suggest additional improvements to alleviate shortcomings
\end{enumerate}

We begin in \cref{sec:history} with a summary of generative AI capabilities. We contextualize the primary efforts that are being made to distinguish between real and synthetic media: namely methods that are detection-based in \cref{sec:detect} and encryption-based in \cref{sec:encrypt}. We believe that encryption-based methods are more future-proof and suggest methods in which they should be incorporated to improve robustness and transparency in \cref{sec:additional_measures}. We conclude, in \cref{sec:conclusion}, by stating the impacts of what it would mean if we can secure our digital future.

\section{Current and Future Capabilities of Generative AI}
\label{sec:history}
The advent of GANs~\cite{goodfellow2014generative} in 2014 has allowed the popularity and photorealism of synthetic media creation, or ``deepfakes", to skyrocket in the recent decade. Some recent applications include face or identity swapping (swapping the face or identity of a target image onto a different source image)~\cite{rossler2019faceforensics++, dale2011video, li2020advancing, nirkin2019fsgan, chen2020simswap, zhu2021one, zhang2021ap, peng2021unified, cao2021unifacegan}, body puppetry (giving a target image the same body posture, gesture, or movements of the source image)~\cite{chan2019everybody, thies2016face2face, thies2019deferred, liu2019neural, doukas2021head2head++, zakharov2019few, wang2019few, gafni2021single, zhang2019faceswapnet, zhang2019one, gu2020flnet, lee2019metapix, sanchez2020recurrent, tripathy2021facegan}, or lip-synching (creating mouth movement in the target image to match any source input audio)~\cite{suwajanakorn2017synthesizing, fried2019text, lahiri2021lipsync3d, zhang2021flow, jamaludin2019you}.

The recent introduction of diffusion-based generative modeling has pushed the boundaries of synthetic media creation even more~\cite{yang2023diffusion}. This class of generative models progressively perturbs input images with noise, and then removes this noise to generate new output images. In the past few years, standard DDPMs~\cite{ho2020denoising, sohl2015deep}, score-based generative models~\cite{song2019generative, song2020improved}, as well as SDE models~\cite{song2021maximum, song2020score, karras2022elucidating} have all arisen as popular formulations with their own set of advantages and disadvantages. Concurrently, further advancements are being made every day on aspects such as more efficient sampling~\cite{chung2022come, zhang2022gddim, zhang2022fast, dockhorn2021score, meng2023distillation, lyu2022accelerating, zheng2022truncated} or better likelihood estimation~\cite{nichol2021improved, NEURIPS2021_b578f2a5, bao2022analytic, huang2021variational, lu2022maximum}, allowing the generative process to become faster and better samplers of the target distribution. The introduction of new architectures like StableDiffusion~\cite{rombach2022high}, which performed generative sampling on a learned latent space as opposed to the image space, led to pushing the boundary even further on the photo-realism and speed of generated images. As these models progressively become better and our access to training data increases in the coming years~\cite{villalobos2022will, ganguli2022predictability}, both the threats and benefits of generative modeling and synthetic media creation will intensify. 

\paragraph{\textbf{The Threats and Benefits of Synthetic Media}}
As our capacity to generate more and more realistic synthetic media increases, so too does our capacity to benefit from the capabilities of generative modeling. Deepfakes have been used for educational purposes, such as the use of a synthetic Barack Obama teaching people about the dangers of deepfakes~\cite{chesney2019deep}. They have also been used commercially in fields like entertainment or marketing~\cite{mirsky2021creation}. Most notably, deepfakes have been used to facilitate social interaction through language barriers or diseases like ALS~\cite{chesney2019deep}.

\begin{table}[t]
  \centering
  \caption{\textbf{A rapid increase in the quality and number of synthetic photos poses several problems to how we can safely navigate digital media.} We identify the primary issues to consider and describe potential solutions in \cref{sec:encrypt}.}
  \begin{tabular}{|p{0.45\linewidth}|p{0.45\linewidth}|}
    \hline
    \textbf{Open Problems} & \textbf{Advocated Solutions} \\
    \hline
    Authenticity of Digital Media (deepfake or real?)  & Verification through cryptography implemented by camera manufacturers. \\
    \hline
    Printout Spoofs & 3D Sensing and light transport \\
    \hline
    Searching for Real Images and Consumer Clarity & Creating new file formats for photos produced by cameras. \\
    \hline
  \end{tabular}
  \label{fig:problems_solutions}
\end{table}

Unfortunately, the advent of generative modeling technology has brought with it a massive host of unwanted threats to security. Generative models’ ability to depict real-world individuals and entities in fabricated settings can have severe consequences for individuals, businesses, or even nations.~\cite{chesney2019deep, dagar2022literature, brooks2021increasing}. In 2021, the DOD published an analysis of the threats of deepfake technology, which included its ability to portray celebrities, business owners, and even political figures in unwanted and potentially harmful manners~\cite{brooks2021increasing}. The article gave numerous examples of how deepfake technology could be used for such unwanted threats as falsifying evidence in criminal court, influencing national elections, or creating non-consensual pornographic material. With the introduction of diffusion models and further improvements in generative modeling in the years that followed this report, the frequency and intensity of threats have only increased. This is why, crucially, methods should be developed to detect the different forms of synthetic media.

\begin{figure*}[t]
    \vspace{-1.5cm}
    \centering
    \includegraphics[width=\textwidth]{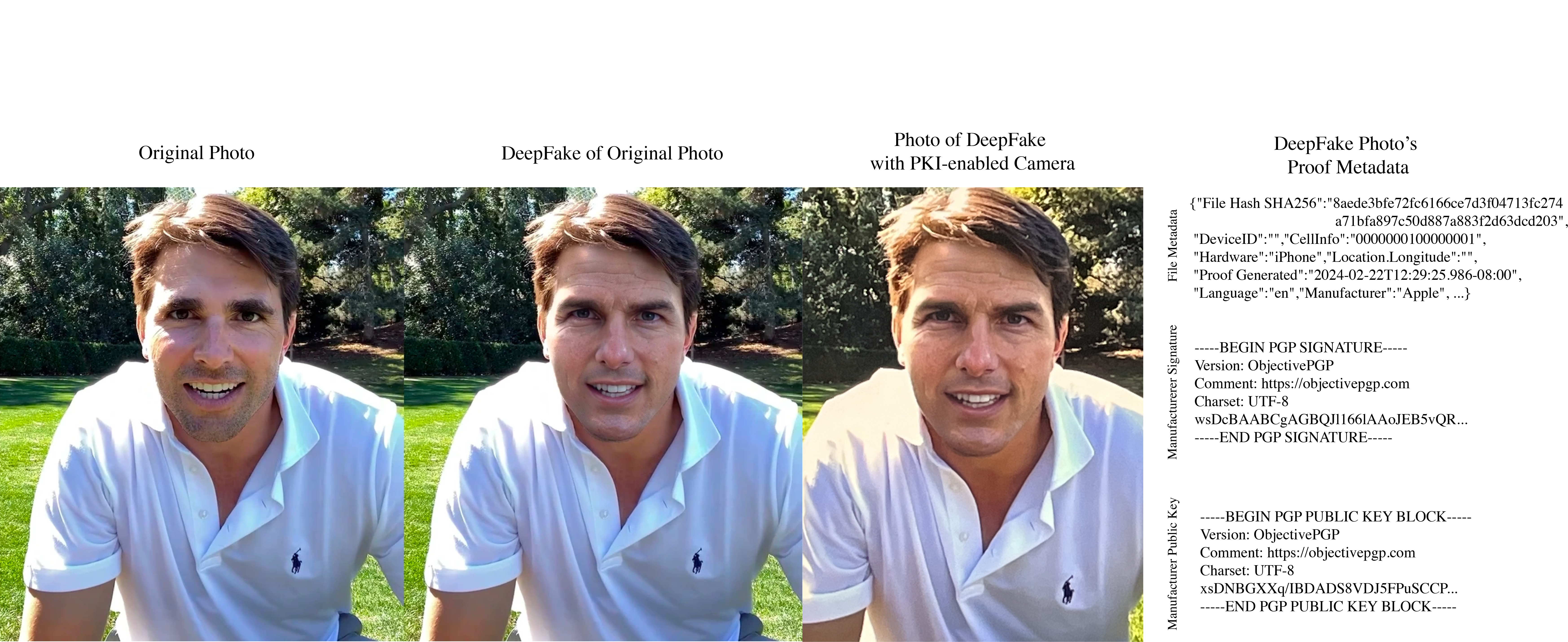}
    \caption{\textbf{C2PA does not guarantee the realness of digital media.} We show an example of taking an image of a deepfake using a camera app implementing C2PA~\cite{proofmode}. The verification of the image will still pass and end-users will be misled into believing that the image is truly real.}
    \label{fig:spoof}
\end{figure*}

\section{Verification through Detection}
\label{sec:detect}
One of the most common existing solutions to the problem of identifying synthetic media is to train machine learning systems designed to identify and detect synthetic images, videos, etc. At a high level, we can define this technique as training a system $\mathcal{D}: I \rightarrow p(R)$, where $\mathcal{D}$ is a detector that takes in an image as input and outputs the probability of it being real.

The design of these detectors can vary a lot, as they can learn to focus on a variety of different cues ranging from systems that look at image statistics and small details to systems that look for biological signs of humans as cues. 

\paragraph{\textbf{DeepFake Detection}} As our generative AI methods improve every day, the systems we make to detect generated media need to improve at a commensurate rate to work properly. Methods like FACTOR rely on similarity scores between ``factual information" given by texts, images, or videos in a pre-trained latent space to obtain a truth rating $p(R)$~\cite{reiss2023detecting}. Other methods utilize popular convolutional models like EfficientNetB4 with attention and ensemble training to get SOTA quality results on detecting manipulated or fake images~\cite{bonettini2021video}. To facilitate the research and discovery of new methods of video deepfake detection, datasets like DFDC (DeepFake Detection Challenge) were produced with over 5K videos across different genders, skin tones, and age groups~\cite{dolhansky2019deepfake}. The current highest performer on the DFDC involves using vision transformers with an EfficientNet B0 encoder~\cite{coccomini2022combining}. The DeepFake detection field~\cite{nirkin2021deepfake, guarnera2020deepfake, zhao2021multi} is rapidly expanding, however, most works are focused on differentiating between fake and real faces and not necessarily fake and real images. Recently, with the increase in the use of diffusion models, several works have aimed at image detection of diffusion-generated content~\cite{corvi2023detection, wang2023dire, lorenz2023detecting}.

\paragraph{\textbf{Arms Race between Generation and Detection}} 
Detection and generation form two components of an adversarial pair. Therefore, each of these can be refined, given the other, to be resilient to the other. This creates an adversarial arms race: the detector can never be guaranteed to be robust to the generator (since the generator may have been refined on the performance of the detector), and hence will need to be refined on all generators. Given this never-ending cycle, detection-based methods can never truly be robust to detecting synthetic media. In the following section, we describe another approach that utilizes encryption to guarantee the authenticity of real digital media. 

\section{Verification through Encryption}
\label{sec:encrypt}
A more robust solution involves cryptographically sealing providence information into a digital media's metadata so that consumers of the content can verify its integrity. The Coalition for Content Provenance and Authenticity (C2PA)~\cite{C2PA, rosenthol2022c2pa} is an organization whose goal is to develop technical standards that can enable such a solution. The C2PA standard proposes a variety of use cases and flexibility, however, we focus on how it can be used to verify whether an image was produced by a camera. In what follows, we detail the salient aspects of how this standard can be used to separate real and synthetic images. 

\paragraph{\textbf{PKI for Camera Generated Digital Media}} The verification of providence is facilitated by incorporating principles from the public key infrastructure (PKI) paradigm that powers SSL/TLS-based internet communications. PKI's goal is to establish a secure transfer of information between two parties. From a cryptography point of view, this is achieved through an interplay between public/private key pairs and a central authority (CA) that all involved parties trust in~\cite{benantar2001internet}. In the case of verifying the authenticity of digital media, we are interested in the following question: how can a user know that a photo was really taken by a camera? This is accomplished in two steps. Before a camera ever takes a photo, the manufacturer of the camera must generate a public and private key pair and register the public key with a central authority. The CA then vets the manufacturer to make sure they are who they say they are and creates a digital signature that the manufacturer can now use. Upon taking a photo, a camera can then sign the data by encrypting a hash~\cite{standard1995federal} of the data using its private key and appending it with the digital signature to the photo's metadata. When the photo is distributed, a consumer can then perform two verification steps. First, the consumer checks that the image is unaltered by decrypting the attached hash using the manufacturer's public key and comparing it with the hash generated by the image data the consumer received. If the hashes are equal, then the image is unaltered. Next, the consumer checks the digital signature with the CA and ensures that the public key is in the trust list of the CA. Then the consumer can be confident that the image was produced by a camera and has been unaltered. 

While PKI-based methods provide a method for verifying that an image came from a camera, it still requires massive coordination between camera manufacturers and is not entirely robust to all forms of spoofing. In the following section, we propose additional modifications to PKI-based methods in the future to become a reliable method for distinguishing between real and synthetic digital media. 

\section{Additional Measures}
\label{sec:additional_measures}

\paragraph{\textbf{Required Steps for Adoption}} The PKI-based approach can only work if it is adopted on a large scale by everyone and the primary burden is on camera manufacturers and the development of a CA that can support such an infrastructure. To achieve a future where it is possible to verify that an image truly came from a camera, several steps must be taken. All camera manufacturers must add the described PKI strategy into their camera pipelines so that each photo can be signed using their private keys and signatures. Secondly, all camera manufacturers must collaborate in creating a unified CA that is devoted to storing a trust list of all public keys that are being used by cameras. The methods for distributing and presenting images must be cognizant of the extra meta-data that digital media may have. These methods must be designed so that they do not accidentally remove the metadata that is used for verification as well as ideally provide their own checks that an image is indeed produced by a camera. The greatest barrier to incorporating this method is in making everyone use it as well as educating media consumers.  

\paragraph{\textbf{Spoofing}} Even when the described approach is implemented correctly, there are still methods by which one can make synthetic images appear real. As mentioned before, the PKI-based strategy only ensures that images were taken by a camera. If the image is altered in any way, the verification process will detect it and flag the image as not real. However, a viewer of an image can still be tricked into believing that a synthetic image is real, barring the possibility that the manufacturer's private keys are leaked. One can still take an image of a synthetic image that is rendered on some surface like a screen or photo using a PKI-enabled camera. We show an example of this process in \cref{fig:spoof}, where we use the app ProofMode~\cite{proofmode}, a program that implements the C2PA standard, as a surrogate for camera hardware. First, a synthetic image is generated and then a picture of it is taken using a camera with PKI signing abilities. Any consumer of the image will be sure that the image was created by a camera and assume that it is real, but it is still a rendition of a synthetically generated image. 

\paragraph{\textbf{Prevention of Spoofing}} The prevention of these types of deceptions is non-trivial. The problem formulation is to create a camera setup that has the following properties:
\begin{enumerate}
    \item It is able to differentiate between natural images of a 3D scene and a 2D scene such as a computer monitor or a flat surface upon which an image can be projected. 
    \item It is not prohibitively expensive. The PKI system will only work with cameras that are robust to spoofing which necessitates a camera setup that is easy to integrate into most systems.
\end{enumerate}
The design of a camera system should optimize its parameters so that it would be difficult for an adversary to simulate an arbitrary image and spoof the system. The design of a camera system can be seen as choosing what regions of the domain, $\theta$, of the plenoptic function~\cite{adelson1991plenoptic}, $\phi(\theta,S)$, should be sampled, where $S$ is a scene. While the plenoptic function's domain is vast, spanning all wavelengths, viewing angles, and other properties of light such as polarization, the cost of sensing such a domain, $c(\theta)$ increases with its size. Informally, the problem can then be viewed as the following optimization problem:
\begin{equation}
    \underset{\theta,D}{\max} \ \ \mathbb{E}_S[\text{log} \ D(\phi(\theta,S)] + \mathbb{E}_G[\text{log} \ 1-D(\phi(\theta,G)] + \beta \ c(\theta),
\label{eq:camera_security}
\end{equation}

where $D$ is a discriminator between real and fake images produced by the plenoptic function to sense a particular domain $\theta$. $S$ and $G$ are random variables for real scenes and adversarially generated scenes that are intended to appear as a real scene. 

Assuming a perfect setup, detection is difficult with only access to one image from a monocular camera due to camera projection of 3D to 2D space, despite prior work on screen recapture detection that typically detects the moir\'e pattern~\cite{choi2017content, li2023two, chen2024cma, mahdian2015identification}. A more robust solution is to detect whether the photo is inherently of a 3D scene. In the case of photos, we are typically restricted to either a single camera or a multi-camera setup. In the case of a single camera, the majority of available techniques rely on structure from motion~\cite{ma2004invitation, agarwal2011building, schonberger2016structure}. Even small jitters caused by a hand holding a camera or smartphone can reveal the 3D nature of a scene ~\cite{yu20143d, im2015high, chugunov2023shakes}. More robust methods are available with a multi-camera setup such as stereo-based methods~\cite{ma2004invitation, hartley2003multiple}. With stereo-based methods, we are able to find correspondences between cameras and triangulate them to find their depth. Here, the detection of an adversarially generated scene can be handled by detecting whether the cameras are viewing a plane. However, even this system is prone to spoofing since attacking this system is tantamount to creating a virtual reality headset for a stereo camera. A more elaborate spoofing system can take advantage of the recently proposed Split-Lohmann multifocal displays~\cite{qin2023split} that would also be able to fool depth-from-defocus~\cite{subbarao1994depth, tang2017depth, watanabe1996real} methods. Fortunately, we have access to many other dimensions of light that are difficult to simulate. For example, smartphones may incorporate new types of sensors such as lidar~\cite{collis1970lidar, li2022deepfusion}, polarization-based sensing~\cite{kalra2020deep, kadambi2015polarized}, or ToF~\cite{freedman2014sra, belhedi2012noise} into their camera systems, all of which are capable of detecting 3D features of a scene. Other solutions may even rely on subtly changing the camera's ISP; for example, spatially altering the Bayer pattern such that the spectral sensitivities of different pixels vary or sparsely sampling a set of pixels to be polarization sensitive could be a low-cost solution to detecting screens. Finding a robust and inexpensive solution to \cref{eq:camera_security} is non-trivial. The better solutions appear to be the ones that take advantage of additional properties of light since it becomes more difficult to pass on these properties to an adversarially generated image, albeit at an increased sensor cost. 

Unfortunately, there exist plenty of real images that are trying to depict a planar surface such as a painting, mural, or a text document without nefarious intent. An anti-spoofing method should not discount such types of media portrayals. A potential solution to this problem is to add a binary label to images that indicate whether an image is primarily a depiction of a 2D or 3D scene. Upon verifying that an image was produced by a camera, viewers of the image should also be told whether the content is of a 2D or 3D nature. It is then up to the viewer to judge whether a particular image is misleading given that information.  

\paragraph{\textbf{New File Types: Faster Searching and Viewer Confidence}} While standards such as those proposed by the C2PA can create a world where images can be verified to be taken by a camera, it does not make the process for searching for real images simple. There are many situations where users may want to search for images that are only taken by a camera. For example, it may be desirable to only train or benchmark certain AI algorithms in the domain of real images. The process of creating such datasets may be time-consuming if it requires a program to search through all images and check their metadata to see if it is real. 

The addition of providence data to a digital media's metadata has several drawbacks. A simple search by file name will not immediately determine if the image has the right type of metadata to prove if it is real and it is simple to remove the providence data from the metadata. File extensions typically serve to indicate the characteristic or intended use of a file. The current file formats for digital media currently do not indicate whether a file format is intended for data that was created by camera hardware. Therefore, a new file format for digital media created by cameras can alleviate these problems. The new file format will include the original data and file format (e.g. .png, .jpg, .mp4) with a new extension (e.g. .png.real, .jpg.real, .mp4.real), but also include a mandatory data field to support PKI-based verification. The programs that read and write such new files can only complete the operation if the mandatory data fields are filled out properly. Moreover, such programs should also incorporate checks to make sure that the image being saved can be verified to be real. With such a new file extension, the set of images that needs to be searched becomes smaller and easier to manage. Moreover, users of the files have more confidence in using such files since the intended use is only for real images. The downside to a new file format is that it requires everyone to add extra support for it. 

\section{What is at stake?}
\label{sec:conclusion}

With generative AI and inevitable progress, the rate, quality, and ease with which synthetic images can be created will be astonishing. It will flood our databases and make it impossible to really know which images are real or fake. Bad actors will be able to create large amounts of highly convincing images and videos to spread disinformation that would be difficult for a layman to disqualify as synthetic or fake. It then becomes imperative that we establish protocols for verifying real digital media before we have reached such a point in the technical advancement of AI. Establishing and implementing such protocols will be a great undertaking that we believe will require careful coordination and co-design of cryptography, camera hardware, and machine learning components. Unfortunately, every day that we have not fully established such protocols, the majority of real images that have been taken today or in the past will not have the potential to be verified as real images. We hope to introduce the broader community to existing methods to distinguish between real and synthetic images, as well as point out the steps required to make these methods robust for the inevitable near future.

\bibliographystyle{acm}
\bibliography{sample-base}

\end{document}